\documentclass[fleqn,10pt]{wlscirep}
\usepackage[utf8]{inputenc}
\usepackage[T1]{fontenc}
\usepackage{subfig}

\title{Protein pathways as a catalyst to directed evolution of the topology of artificial neural networks}

\author[1,*]{Oscar Lao}
\author[2]{Konstantinos Zacharopoulos}
\author[4]{Apostolos Fournaris}
\author[5,6]{Rossano Schifanella}
\author[3,*]{Ioannis Arapakis}

\affil[1]{IBE, Institute of Evolutionary Biology (CSIC-Universitat Pompeu Fabra), Barcelona, 08003, Spain}
\affil[2]{Technical University of Crete, Chania, 731 00, Greece}
\affil[3]{Telef\'{o}nica Innovaci\'{o}n Digital, Scientific Research, Barcelona, 08019, Spain}
\affil[4]{ATHENA, Industrial Systems Institute, Patras, 265 04, Greece}
\affil[5]{Computer Science Department, University of Turin, Turin, 10149, Italy}
\affil[6]{ISI Foundation, Turin, 10126, Italy}
\affil[*]{corresponding.author@email.example}


\begin{abstract}

In the present article, we propose a paradigm shift on evolving Artificial Neural Networks (ANNs) towards a new bio-inspired design that is grounded on the structural properties, interactions, and dynamics of protein networks (PNs): the \textbf{Artificial Protein Network (APN)}. This introduces several advantages previously unrealized by state-of-the-art approaches in NE: (1) We can draw inspiration from how nature, thanks to millions of years of evolution, efficiently encodes protein interactions in the DNA to translate our APN to silicon DNA. This helps bridge the gap between \textit{syntax} and \textit{semantics} observed in current NE approaches. (2) We can learn from how nature builds networks in our genes, allowing us to design new and smarter networks through EA evolution. (3) We can perform EA crossover/mutation operations and evolution steps, replicating the operations observed in nature \textit{directly} on the genotype of networks, thus exploring and exploiting the phenotypic space, such that we avoid getting trapped in sub-optimal solutions. (4) Our novel definition of APN opens new ways to leverage our knowledge about different living things and processes from biology. (5) Using biologically inspired encodings, we can model more complex demographic and ecological relationships (e.g., virus-host or predator-prey interactions), allowing us to optimise for multiple, often conflicting objectives.

\end{abstract}
\begin{document}

\flushbottom
\maketitle
\thispagestyle{empty}

\noindent Please note: Abbreviations should be introduced at the first mention in the main text --- no abbreviations lists. Suggested structure of main text (not enforced) is provided below.

\section*{Introduction}

Deep Learning (DL) is inspired by how nature engineered our brain for recognising highly complex patterns, storing information, integrating inputs, and producing a thought or an action. During the last decade, DL has led to groundbreaking results in many fields, pointing out the potential of this bioinspired paradigm. Nevertheless, crafting a successful neural architecture (i.e., identifying hyperparameters such as how many neurons, how neurons are connected among layers, or which activation functions are used) of a DL model is a nondeterministic polynomial (NP) time-complete issue~\cite{SCHMIDHUBER201585}). Historically, as the theory beneath why particular de novo neural architectures should work for a given problem is still in its infancy, in practice, engineering an optimal architecture depends heavily on the researchers' pre-existing knowledge, experience, or even intuition. Innovation in ANN architectures has usually been driven by emulating the functionality of naturally evolved circuits. For example, one prominent biological inspiration for convolutional neural networks is the structure and operation of the visual cortex, which is responsible for visual information processing~\cite{LeCun2015}. The need for substantial expertise poses a challenge for beginners, who can apply underperformed models to their specific requirements. Additionally, individuals' existing knowledge and entrenched thinking paradigms may potentially constrain the exploration and discovery of new neural architectures to a certain extent. In line with this perspective, current DL models are prone to deception, misclassification, and getting trapped in local minima~\cite{Ahmed2023, Shrestha2019}. 
Within the DL field, Neural Architecture Search (NAS) addresses the challenge of autonomously designing a neural architecture that attains optimal performance with constrained computing resources while requiring minimal human intervention~\cite{Ren2021}. NAS typically starts with a predefined set of operations and employs a controller to generate numerous candidate neural architectures within the search space defined by these operations. Subsequently, these candidate architectures undergo training on a designated training set and are ranked based on their accuracy on a validation set. The ranking information of these candidates serves as feedback to adjust the search strategy, facilitating the generation of a new set of candidate neural architectures~\cite{Ren2021}. Different approaches based on Reinforcement Learning~\cite{Zoph2016}, Bayesian Optimization~\cite{White2019}, or Monte Carlo tree search~\cite{Su2021}, among many others, have been proposed for DL architecture optimization~\cite{Ren2021}. Nevertheless, evolutionary approaches are particularly interesting since NAS strategies are the closest to mimicking the natural process that engineered the brain, and they offer substantial flexibility in their implementation and application. 
Biological evolution acts on the \textit{phenotype} (i.e., the set of observable characteristics of an organism), which is how the information stored in DNA (i.e., \textit{genotype}) is expressed as a product of interaction with the environment. The evolutionary process unfolds within a finite population of systems, which engage in self-replication with random errors and reproduce at specific deterministic rates based on their interaction with the environment. This interplay gives rise to competition among systems within the population, each striving to produce more copies over time. Viewed geometrically, the entire process can be seen as projecting each individual into the phenotypic space for a fitness assessment. The creation of new individuals involves navigating the phenotypic space between selected parents. This ongoing process systematically explores the phenotypic space, adapting optimally to the environment. This competitive dynamic ultimately drives the adaptation of the systems to their environment. There is no design process in this context, and many optimizations can be achieved fortuitously. 

\section*{Evolution as a powerful meta-optimisation algorithm}

\begin{figure}[ht]
\centering
\includegraphics[width=\linewidth]{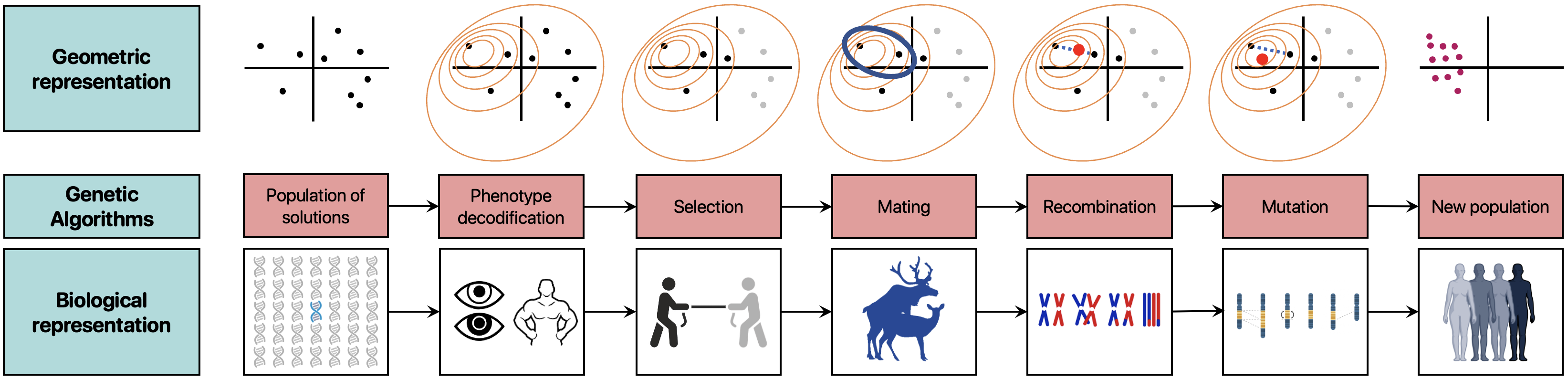}
\caption{Evolution serves as a metaheuristic for optimization. Genetic diversity within a population leads to varied phenotypes influenced by environmental factors. Through reproduction, genetic material merges, creating novel phenotypes. This process involves navigating phenotypic space for fitness assessment and systematically exploring and adapting to the environment. A genetic algorithm mimics this process.}
\label{fig:evolution}
\end{figure}

The universality of evolution to solve highly complex problems has been acknowledged for a long time ~\cite{Brabazon2015}, inspiring a broad field of metaheuristic algorithms called Evolutionary Algorithms (EAs). Some of the desirable properties of EAs are (1) global optimization, (2) exploration of diverse search spaces, (3) handling complex functions, (4) independence from derivative calculations, (5) ease of parallelization, (6) robustness against noise, and (7) compatibility with other algorithms outweigh~\cite{Brabazon2015}.
EAs work with \textit{encoded} information into chromosomes-like structures. The fitness of these encodings is evaluated by \textit{decoding} them into solutions for specific problems and measuring how well they match expected outcomes. The language used in EAs, like in Genetic Algorithms (GA), mirrors biological terms. GA involves selecting high-fitness solutions, resembling ``survival of the fittest'', and using operations like crossover and mutation to improve solutions (Fig. \ref{fig:evolution}). 

In recent decades, EAs have effectively solved the complex optimization of poorly understood problems in various fields~\cite{Brabazon2015}. Recognizing that biological neural networks (i.e., brains) result from a product of evolution over millions of years~\cite{Kristan2016} (Fig. \ref{fig:brain-creation}) gave rise to the field of Neuroevolution (NE) to automatically design Artificial Neural Networks (ANNs). NE is used for tasks like estimating weights, pruning nodes, adjusting hyperparameters, or generating architecture and parameters for ANNs~\cite{lanham2023}. Successful applications often involve compact ANNs with hundreds or thousands of connections~\cite{Stanley2019}. Nevertheless, NE does not quite match the remarkable efficiency observed in nature. This is partly attributed to NE being an extension of evolutionary algorithms (EAs), which means it encounters the same potential drawbacks as other EA applications. EAs limitations include (1) the need for encoding problems, (2) uncertainty in consistently providing optimal solutions within a set timeframe, (3) potential parameter tuning, and (4) the demand for significant computational resources. While certain drawbacks, like the requirement for computational resources, are diminishing rapidly in NE due to advancements in hardware, others still pose more elusive challenges to be resolved in NE. In particular, one of the main issues compromising the performance of NE implementations is related to how a neural network is encoded in a chromosome. NE suffers from a gap between \textit{syntax} (how elements are arranged in the chromosome) and \textit{semantics} (their function in the network). This gap can lead to different syntax conveying the same meaning or similar functions having entirely different syntax. Ideally, the syntactic encoding of any neural network within a chromosome should ensure that when two chromosomes combine to generate a new offspring by crossover and mutation, the output will be \textit{semantically preserved}, thereby giving the evolutionary process a gradient it can capitalize on. The evolutionary process is hindered when this is not achieved, resulting in expensive, poorly performing neural networks.
Current NE strategies use various encoding methods to capture evolving network characteristics. Direct encoding methods (i.e., directly coding neurons, connections, and weights in a string) use matrices, enumeration, or genetic trees, allowing crossover based on node order~\cite{Fekia2011}. These methods do not align with natural phenomena, limiting their ability to mimic biological evolution's nuances. This compromises adaptability and diversity, hindering the potential for effective ANNs. Encoding strategies assuming fixed graph structures also prove incompatible in NE, where innovative graph features are common. Moreover, direct encoding approaches face challenges with larger, highly connected graphs. Alternative indirect encodings mimicking the natural encoding have been proposed~\cite{Hara2003,Bordin2023}, e.g., artificial ontogeny or artificial embryogeny to model embryonic development. However, there is flexibility in not strictly simulating biological development, and some approaches make the design of these structures an intuitive and trial-and-error effort~\cite{Matos2009}.

\subsection*{Nature as inspiration for encoding-decoding Neural Networks in NE}

Nature has developed effective ways of encoding \textit{phenotypic characteristics} in {chromosomes}, which introduces many advantages that make evolution a very efficient \textit{meta-optimisation} mechanism. These encoding properties prevent organisms from converging to sub-optimal environmental adaptations or less-than-perfect designs too quickly. In biological systems, information defining a biological neural network is efficiently encoded~\cite{Salzberg2018O,HerculanoHouzel2009} in the DNA, thanks to billions of years of evolution on encoding and decoding information to thoroughly explore the phenotypic space~\cite{Hara2003}. The biological transition from syntax to semantics involves a complex decoding process. This efficiency in compressing information likely underwent selective shaping throughout evolution, so biological organisms could endure recombination and mutation to improve adaptation towards the environment while maintaining optimal compression rates. The decoding process includes a set of complex molecular and developmental steps (Fig. \ref{fig:brain-creation}). Moreover, biological neurons can adapt and evolve based on environmental factors, resulting in phenotypic changes from the initial cell to the adult form.

\begin{figure}[ht]
\centering
\includegraphics[width=\linewidth]{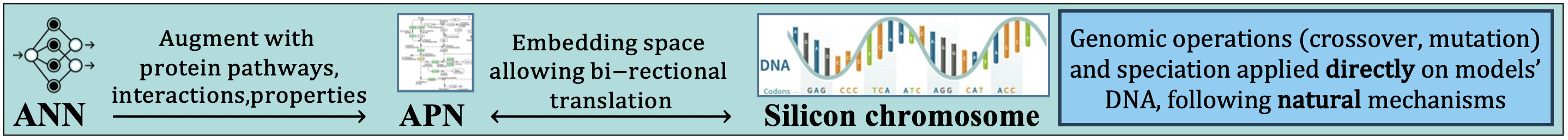}
\caption{Example of an encoding-decoding pathway of Artificial Protein Network (APN) to silicon chromosome}
\label{fig:intro-fig}
\end{figure}

Following the natural computing principles, one could draw inspiration from biological systems near DNA and network systems that resemble how information is processed in biological neural networks. Here, we propose to develop a new ANN design and training paradigm by drawing inspiration from how cells encode in the DNA signalling processes (Fig.~\ref{figure5_a}) that allows AI to evolve and adapt holistically and coherently. As DNA repositories, cells process information similarly to the brain, aligning with broader information processing principles. We can draw parallels with the cell to the engineering and design principles seen in the brain~\cite{Lim2014} (Fig. \ref{fig:brain-creation}). Within this analogy, Protein-Protein Interactions (PPI) can manifest as direct (physical) or indirect (functional) interactions and may be directed (i.e., protein A modifies the performance of B) or undirected (i.e., protein A and B contact each other)~\cite{Tang2022} that would resemble the connections observed between neurons in the brain. 

\begin{figure}[ht]
\centering
\includegraphics[width=0.9\linewidth]{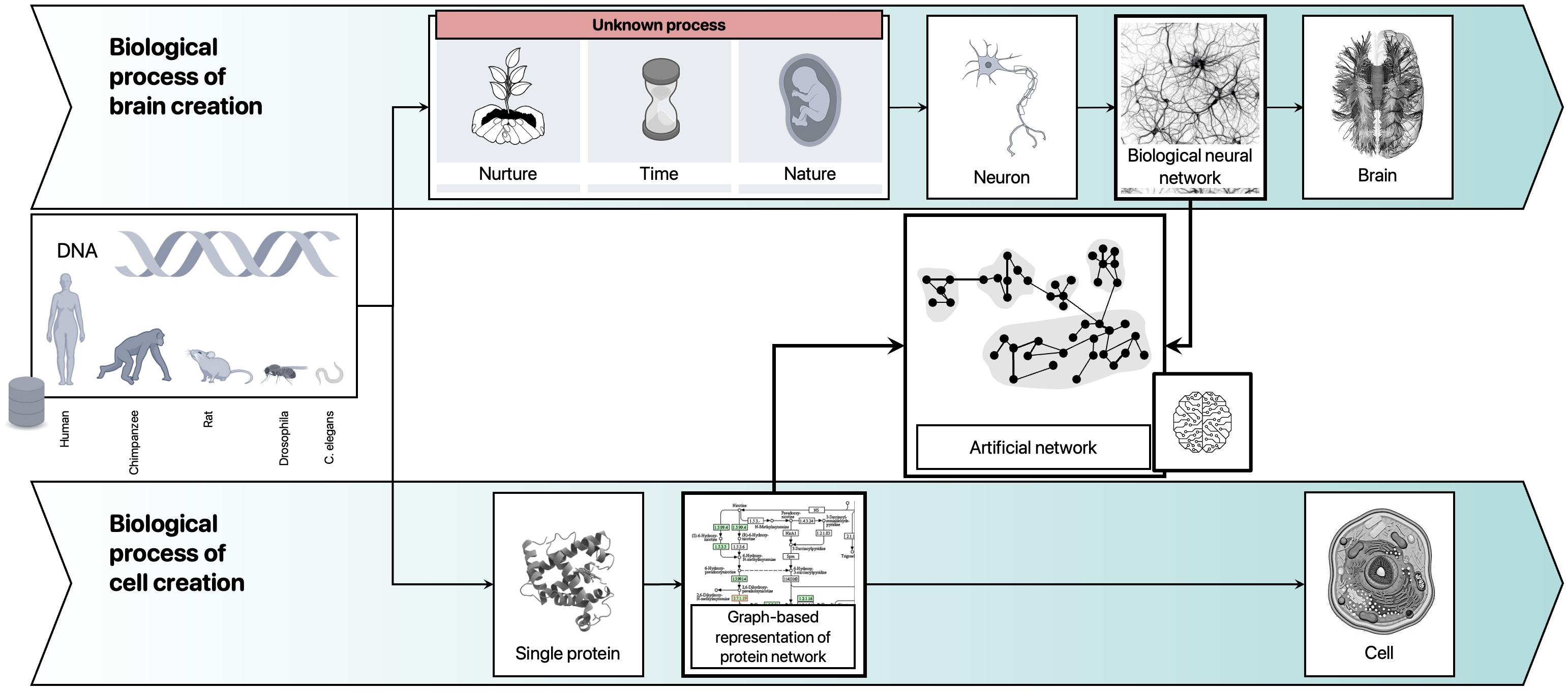}
\caption{Biological similarities of PNs and NNs. The genetic blueprint of a species contains both NNs and PNs. Neurons are at the core of NNs, forming neural tissue up to the brain. Decoding NNs from DNA is difficult as NNs are shaped by environment and development. Directed signed PNs use proteins as basic units, with functions directly dictated by DNA, that enable PPIs and forming PNs that contribute to cell functionality.}
\label{fig:brain-creation}
\end{figure}

\begin{figure}[hb!]
 \centering
	\captionsetup[subfloat]%{captionskip=-5pt,nearskip=0pt,farskip=0pt}
	{}
	\begin{minipage}[c]{0.5\linewidth}
	\centering
    	\subfloat[]{%
    	\includegraphics[trim={0 0 0 0}, clip, scale=0.17]{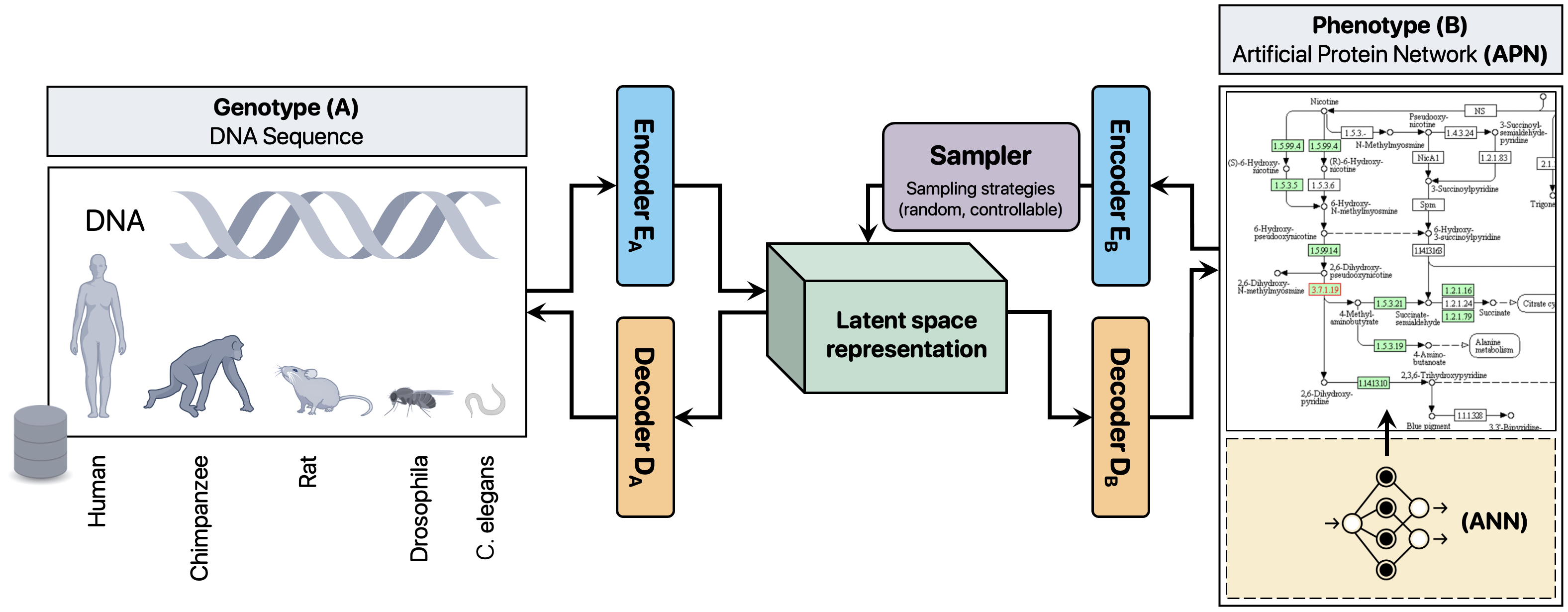}
    	\label{figure5_a}
    	}
	\end{minipage}
	\hspace{8mm}
	\begin{minipage}[c]{0.22\linewidth}
	\centering
    	\subfloat[]{%
    	\includegraphics[trim={0 0 0 0}, clip, scale=0.12]{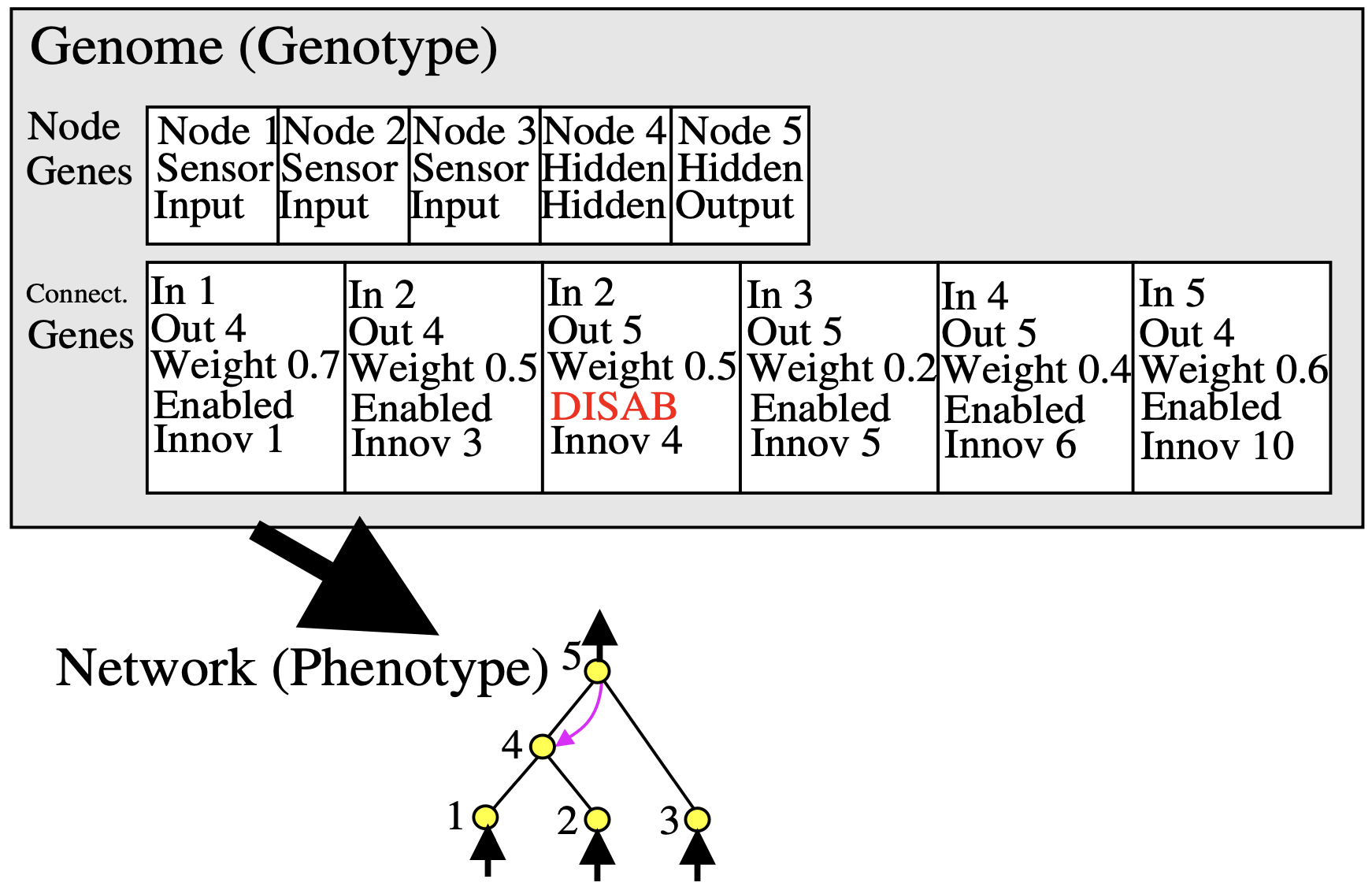}
    	\label{figure5_b}
    	}
	\end{minipage}
	% \hspace{0.5mm}
	\begin{minipage}[c]{0.22\linewidth}
	\centering
    	\subfloat[]{%
    	\includegraphics[trim={0 0 0 0}, clip, scale=0.12]{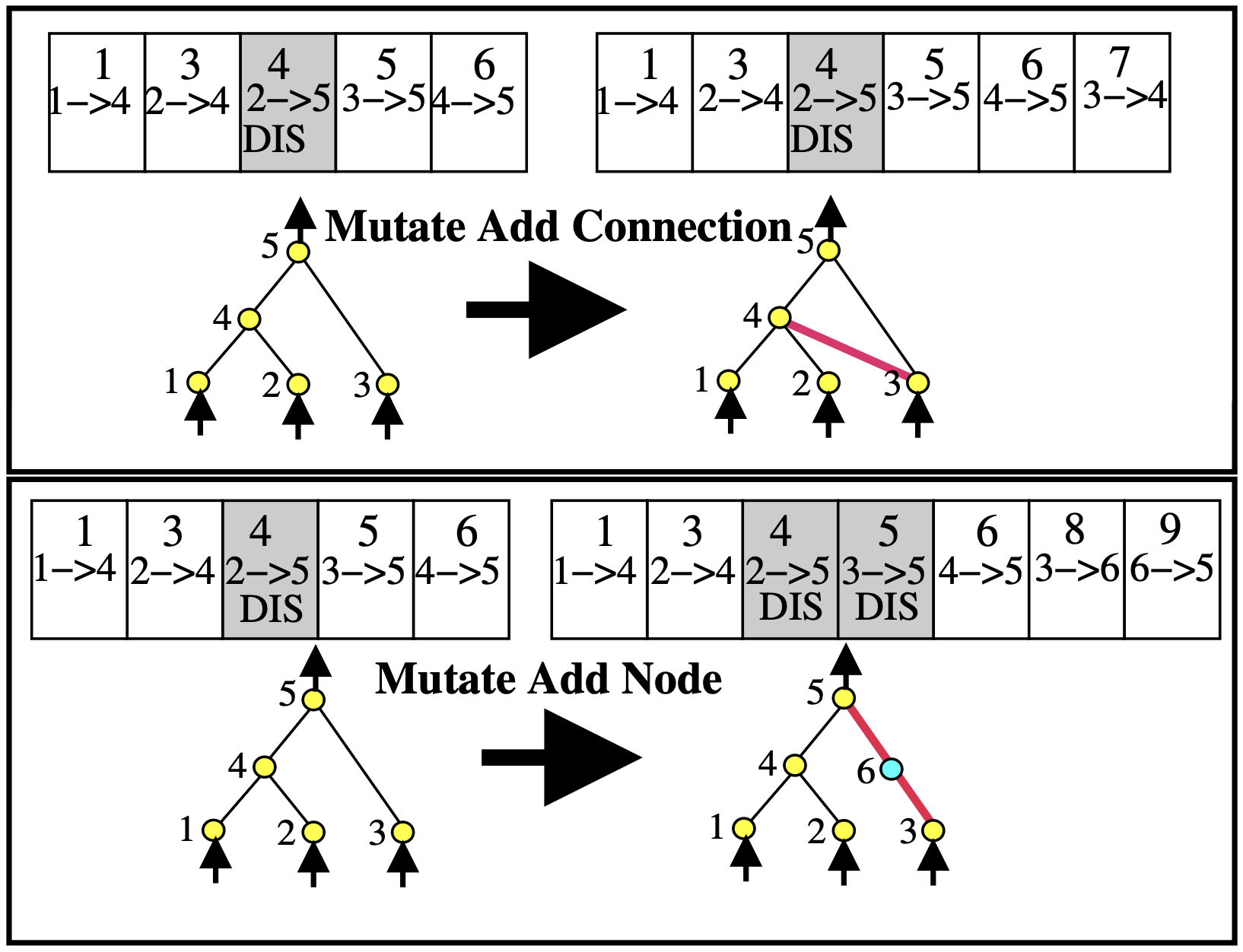}
    	\label{figure5_c}
    	}
	\end{minipage}
\caption{Left side: (a) A DL model for emulating the biological non-linear decoding of the genetic information present in the DNA to generate a network. Right side: (b) Genotype to phenotype mapping and (c) structural mutilation examples, as implemented in NEAT~\cite{Kenneth2002}}
\label{fig:encoder-decoder}
\end{figure}

In this context, an artificial protein network (APN) would be the product of encoding an ANN to a (directed, weighted) graph architecture that encapsulates rules on how proteins exchange information through pathways in a biological cell. In essence, an APN would implement an ANN and augment it with the properties of a PN, where proteins act as node-receptors, and the connections-interactions between them are represented as edges (Fig. \ref{fig:fig-ANN-vs-APN}). 

\begin{figure}[ht]
\centering
\includegraphics[width=0.9\linewidth]{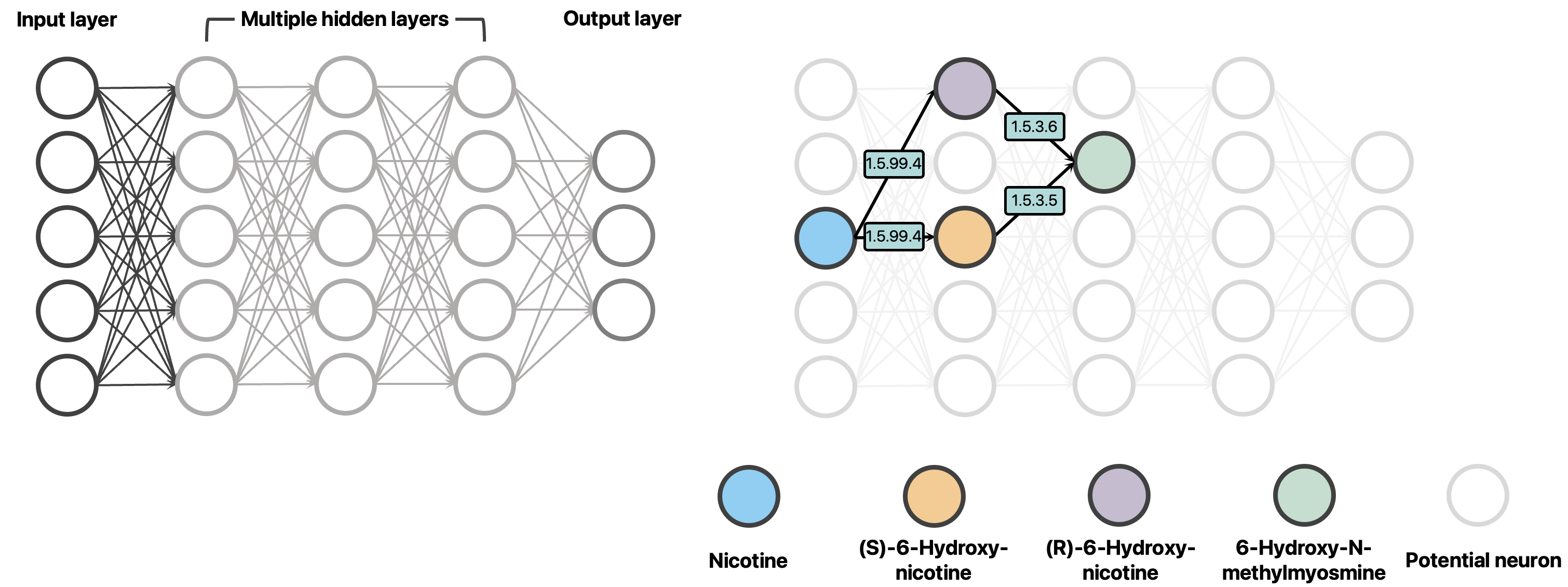}
\caption{An artificial neural network is mirrored by particular connections between proteins.}
\label{fig:ann-vs-apn}
\end{figure}

Therefore, by designating each neuron as a specific protein, we inherit its properties, allowing a new ANN design/training paradigm, aligned with biological evolution principles for holistic adaptation.
In the proposed system, the syntactic vs. semantics gap that limits the application of biological recombination and mutation in NE (Fig.~\ref{figure5_b}-\ref{figure5_c}) is bridged by learning how biological sequences encoding biological proteins lead to protein networks through leveraging generative AI, and in particular Transformer architectures, to achieve a Genotype-to-Phenotype mapping, and vice versa. Transformer architectures have been successfully applied to create protein language models, i.e., protein Learning Models (pLMs), that have been transformative in predicting protein structures~\cite{Bordin2023}. DL transformers, such as AlphaFold 2~\cite{Jumper2021} and AlphaFold 3~\cite{Abramson2024}, effectively handle graph-based and sequence-based structures, capturing local and global dependencies. Adaptability allows transformers to model highly complex PPIs accurately, especially long-range interactions ~\cite{Jumper2021, Gao2022, Jinhyuk2019, Mirdita2022}. Leveraging these features, our goal is to decode the syntax of PNs and embed it into ANN architectures, where each neuron will represent a specific protein along with its unique properties.
More specifically, we propose that given a database of DNA sequences or protein sequences (Genotype A), sampled from various biological species, and a curated collection of cell-signalling networks and associated protein features (Phenotype B), we can train Transformers (Fig.~\ref{figure5_a}) to acquire a latent space representation that maps an APN to a silicon chromosome. We propose using protein pathways and genomic data from numerous public databases~\cite{Tieri2013} to curate datasets of biological pathways for DNA multialignment.
Essentially, the completion of this step will allow the bi-directional translation between an APN architecture (i.e. Phenotype B) and its DNA sequence (i.e. Genotype A), by means of projecting both onto the same latent space, hence, realising an encoding that is grounded in biology. This integration assimilates each protein's connection rules and limitations, ultimately leading to an APN and, by analogy, to an ANN (Fig. \ref{fig:brain-creation}).

In the conventional NE approach, automatically initialising a population of NNs involves randomly generating their architectures and encoding them directly into silicon genomic material. In the APN strategy, de novo generated architectures must be bio-inspired to allow the trained transformer to generate a chromosome-encoded representation of the APN and vice versa (Fig. \ref{fig:encoder-decoder}). Alternatively, more efficient initializations can be achieved by leveraging prior knowledge of existing neural networks --- i.e., understanding why a particular network architecture works for a given task. However, these practices present a significant challenge in the proposed DNA-encoder-to-APN-decoder (and vice versa) approach, as the randomly generated network must be translated into silicon protein sequences. Generating a biologically plausible protein network de novo or by taking inspiration from known non-protein related architectures involves graph matching that is generally known to be NP-hard~\cite{Loiola2007}, which is addressed mainly through approximate techniques and often leads to inexact solutions. Moreover, real-world graphs --- like complex protein interaction networks --- often show complex non-linear patterns that are hard to describe with predefined metrics~\cite{Albert2002, Leskovec2010, Minsu2010}. We aim to create a deep generative framework to generate seed APNs for the evolutionary process. In particular, deep graph generative models, using neural architectures, capture complex dependencies and patterns in real-world networks, allowing end-to-end learning. A second approach for initialising a population of architectures can be achieved through innovative graph-matching techniques. The \textit{inexact} or \textit{error-tolerant graph matching} variant aims to solve the problem of finding node and edge correspondences over two or multiple graphs of different sizes. It incorporates both node-wise unary similarity and edgewise~\cite{Minsu2010} (or even higher-order~\cite{Yan2015, Ngoc2015}) similarity to establish a matching to maximise the similarity between the matched graphs. Here, a graph embedding solution using GNN architectures is proposed~\cite{Cai2017}, preserving properties of interest for efficient matching in the projected space.

The APN has the advantage that, due to its proximity to the DNA source (through the protein sequence information), it leverages the wealth of existing biological databases to learn how to translate between an APN network and DNA. This further enables the application of biological-like recombination and mutation operations on the AI models' DNA to evolve new synthetic genes. In NE, traditional crossover and mutation operators are inspired by biological recombination and mutation but can be disruptive due to differences in silicon and biological chromosome encoding. Challenges include functional interdependence among connection weights~\cite{Qiao2023} and the permutation problem, where diverse configurations yield identical functionality through the reordering of neurons, as exemplified in a fully connected layer~\cite{Gao2022}. Advanced algorithms (e.g., NEAT~\cite{Kenneth2002}) introduce ``speciation'' to capture network semantics. In natural systems, recombination and mutation are intertwined, evolving to enhance species long-term survival and adaptability~\cite{Lynch2016}. Nevertheless, by using a closer model of biological DNA, methods from the field of phylogenetics can be used to define degrees of homology between genomes and genes, closely mimicking the process observed in life for matching two genomes. Crossover breakpoints can be introduced into the sequences based on evolving motifs of genetic material alongside the silicon proteins responsible for recognizing these motifs. Therefore, recombination, driven by sequence homology, could introduce new genomic material and innovative possibilities like gene duplications, deletions, chimeras, etc., as observed in biological organisms ~\cite{Kaessmann2010}. Mutation strategies can be inspired by genetic variation databases, mimicking natural processes like single amino acid changes, de novo gene creation, deletions, or fusions, among others.

\begin{figure}[ht]
\centering
\includegraphics[width=\linewidth]{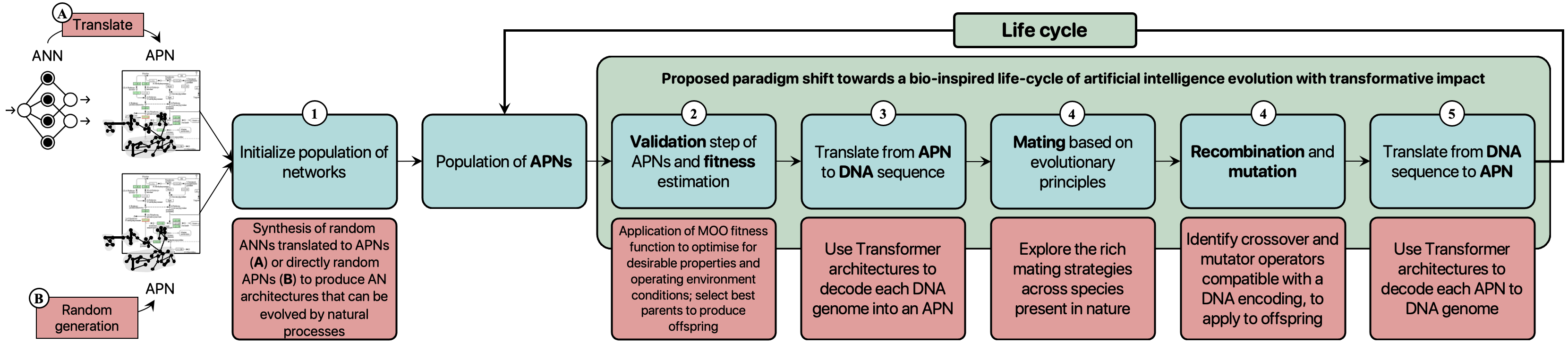}
\caption{Overview of disruptive long-term vision and proposed life cycle of AI evolution.}
\label{fig:overview}
\end{figure}

Implementing the APN approach (Fig. \ref{fig:overview}) would require: (1) Define the ecosystem and demography of the population of networks, (2) Start a population of networks by initialising network structures and mapping them to PNs. Once the protein-like chromosome is obtained, (3) identify the most successful networks, (4) translate the APN networks to DNA, and (5) use them to create new solutions by applying biologically inspired mutations (altering amino acids, duplicating genes, creating chimeric proteins by fusing two different genes, etc.) and recombination (guided by protein structure similarity). New solutions are then (6) translated again to APNs, thus finishing a generation of the life cycle.

\subsection*{Enhancing operational adaptability of Artificial Neural Networks}

The APN framework, similarly to other NE algorithms, is quite adaptable. For example, the APN-cellular analogy provides the background for emulating processes related to cellular co-evolution~\cite{bucciarelli2022}, which could be used by reverse engineering to model complex neural networks adapting to dynamic environments with constraints such as low power, minimal memory, or high speed, by incrementally downsizing structural complexity, mirroring gene expression regulation in cells. This is particularly relevant in the DL field, as current ANN adaptability by techniques like quantization, pruning, and sparsity enhancement~\cite{Lazzaro2023} lacks automation, resulting in performance overhead. Some prior work incorporated temporal dynamics into the deep NNs with promising results~\cite{Lechner2020}. 

A potential field to apply the proposed APN framework would be mobile communication systems, which provide global-scale infrastructure for mobile users worldwide, as that field aims to optimise paths, reduce latencies, and minimise infrastructure involvement for sustainability. More specifically, as we transition between mobile generations, e.g., from 4G to 5G, and cloud– to-edge, AI/ML technologies can act as optimization enablers and realise a transformative impact. However, they can also introduce substantial energy consumption demands. While 5G is more energy-efficient than 4G in transmission, it could significantly increase energy demand in data centers by up to 3.8 terawatt hours (TWh)~\cite{Hoefer2024} by 2025. This considerably impacts distributed computing at large, i.e., for edge computing, mixed virtualized/physical networking, and data centers together. Moreover, AI/ML algorithms involve massive data transfers and computational intensity, contributing to energy consumption~\cite{Patterson2021}. Even with available computational resources, model training incurs environmental costs due to prolonged hardware usage. Regardless of the energy production source, high energy demands remain a concern due to limitations in renewable energy infrastructure~\cite{Strubell2019}. The APN framework offers the prospect of delivering nimble AI/ML architectures capable of running energy-efficient and energy-adaptable APN models. These adaptable model summaries will incur a small carbon footprint during training/execution and reduce data processed per edge device. Operating within energy and hardware constraints across the network infrastructure, the APN architectures may help balance environmental sustainability and operational accuracy, aiding the transition to a low-carbon economy~\cite{GreeningICT}.

Additionally, in the cyber domain, ANNs face adversarial and privacy attacks. Securing cyberspace is a top priority in all public and private sectors, highlighting the demand for cybersecurity professionals\cite{CyberSeek2024}. To address the shortage and avoid the economic impact of cyberattacks, companies around the globe are turning to advanced AI solutions for cyber threat detection, creating a market projected to grow at a compound annual rate of 21.9\% from 2023 to 2028, reaching a revenue of \$60.6 billion by 2028~\cite{AICybersecurity}. However, the swift progress of AI technology has led to the emergence of AI-assisted cybercriminals and a novel cyber threat landscape that is far more dangerous compared to the pre-AI era. AI-driven malware adapts to target environments, making detection challenging (e.g., social engineering attacks using AI, such as deepfake-generated phishing messages). The cybercrime economic impact is substantial, with a projected global loss of \$600 billion annually, nearly one percent of global GDP~\cite{Lewis2018}, rising to \$10.5 trillion for the business sector by 2025~\cite{economicimpact}. Some preliminary research models cyberspace as an ecosystem of cyber organisms~\cite{Lippert2021} that co-evolve by predation, parasitism, symbiosis, and cooperation~\cite{Bacar2011}. Bioinspired cyber defense systems~\cite{Guthikonda2017} aims to reduce reliance on sample analysis but often function similarly to traditionally trained ANNs with data-intensive training. In the proposed APN framework, we can expand on the cybersecurity arms race between attackers and defenders as a practical paradigm of co-evolution, where ecological models can be applied to enhance APNs with cyber-attack resistance and self-protection characteristics, achieving significant reductions to the cyberattacks' economic losses.

\subsection*{Closing Remarks}

Neuroevolution has yet to fully replicate the complexity of neural structures observed in biological systems, a challenge that stems significantly from the NP-hard problem of finding a suitable encoding-decoding of the network into genetic material. This limitation often leads to the evolutionary process being entrapped in local optima, thereby stifling progress in the domains of NE and EAs. In this paper, we propose for the first time a paradigm shift in the definition of ANNs and present a novel bio-inspired design grounded on the structural properties, interactions, and dynamics of PNs - the Artificial Protein Network.

We argue that the proposed APN framework can introduce several previously unrealized advantages by other state-of-the-art NE approaches. Firstly, it allows us to draw inspiration from nature, which has efficiently encoded protein interactions in DNA over millions of years, to translate our APN onto silicon DNA, thereby bridging the syntax-semantics gap observed in current NE approaches. Secondly, we can learn from how nature constructs networks in our genes, enabling the design of smarter and more sophisticated networks through EA evolution. Additionally, the proposed APN framework will allow us to replicate EA crossover/mutation operations and evolution steps directly on the genotype of networks, thus exploring and exploiting the phenotypic space and preventing us from getting trapped in sub-optimal solutions. Furthermore, this novel definition of APN paves the way to leverage our knowledge about various biological processes and organisms. Last, the use of biologically inspired encodings permits us to model more complex demographic and ecological relationships, such as virus-host or predator-prey interactions, which will enable us to optimize for multiple, and often conflicting, objectives.

In light of our unique positioning with respect to biological evolution and the open problems that impede progress in NE and EAs, we anticipate that this paradigm shift will have substantial implications for numerous technological domains, including cybersecurity and telecommunications.

% ----------------------------------------------------------------------------

\bibliography{sample}

\begin{thebibliography}{10}
\urlstyle{rm}
\expandafter\ifx\csname url\endcsname\relax
  \def\url#1{\texttt{#1}}\fi
\expandafter\ifx\csname urlprefix\endcsname\relax\def\urlprefix{URL }\fi
\expandafter\ifx\csname doiprefix\endcsname\relax\def\doiprefix{DOI: }\fi
\providecommand{\bibinfo}[2]{#2}
\providecommand{\eprint}[2][]{\url{#2}}

\bibitem{SCHMIDHUBER201585}
\bibinfo{author}{Schmidhuber, J.}
\newblock \bibinfo{journal}{\bibinfo{title}{Deep learning in neural networks:
  An overview}}.
\newblock {\emph{\JournalTitle{Neural Networks}}}
  \textbf{\bibinfo{volume}{61}}, \bibinfo{pages}{85--117},
  \doiprefix\url{https://doi.org/10.1016/j.neunet.2014.09.003}
  (\bibinfo{year}{2015}).

\bibitem{LeCun2015}
\bibinfo{author}{LeCun, Y.}, \bibinfo{author}{Bengio, Y.} \&
  \bibinfo{author}{Hinton, G.}
\newblock \bibinfo{journal}{\bibinfo{title}{Deep learning}}.
\newblock {\emph{\JournalTitle{Nature}}} \textbf{\bibinfo{volume}{521}},
  \bibinfo{pages}{436--44}, \doiprefix\url{10.1038/nature14539}
  (\bibinfo{year}{2015}).

\bibitem{Ahmed2023}
\bibinfo{author}{Ahmed, S.} \emph{et~al.}
\newblock \bibinfo{journal}{\bibinfo{title}{Deep learning modelling techniques:
  current progress, applications, advantages, and challenges}}.
\newblock {\emph{\JournalTitle{Artificial Intelligence Review}}}
  \textbf{\bibinfo{volume}{56}}, \doiprefix\url{10.1007/s10462-023-10466-8}
  (\bibinfo{year}{2023}).

\bibitem{Shrestha2019}
\bibinfo{author}{Shrestha, A.} \& \bibinfo{author}{Mahmood, A.}
\newblock \bibinfo{journal}{\bibinfo{title}{Review of deep learning algorithms
  and architectures}}.
\newblock {\emph{\JournalTitle{IEEE Access}}} \textbf{\bibinfo{volume}{7}},
  \bibinfo{pages}{53040--53065}, \doiprefix\url{10.1109/ACCESS.2019.2912200}
  (\bibinfo{year}{2019}).

\bibitem{Ren2021}
\bibinfo{author}{Ren, P.} \emph{et~al.}
\newblock \bibinfo{journal}{\bibinfo{title}{A comprehensive survey of neural
  architecture search: Challenges and solutions}}.
\newblock {\emph{\JournalTitle{ACM Comput. Surv.}}}
  \textbf{\bibinfo{volume}{54}}, \doiprefix\url{10.1145/3447582}
  (\bibinfo{year}{2021}).

\bibitem{Zoph2016}
\bibinfo{author}{Zoph, B.} \& \bibinfo{author}{Le, Q.~V.}
\newblock \bibinfo{journal}{\bibinfo{title}{Neural architecture search with
  reinforcement learning}}.
\newblock {\emph{\JournalTitle{CoRR}}}
  \textbf{\bibinfo{volume}{abs/1611.01578}} (\bibinfo{year}{2016}).
\newblock \eprint{1611.01578}.

\bibitem{White2019}
\bibinfo{author}{White, C.}, \bibinfo{author}{Neiswanger, W.} \&
  \bibinfo{author}{Savani, Y.}
\newblock \bibinfo{journal}{\bibinfo{title}{{BANANAS:} bayesian optimization
  with neural architectures for neural architecture search}}.
\newblock {\emph{\JournalTitle{CoRR}}}
  \textbf{\bibinfo{volume}{abs/1910.11858}} (\bibinfo{year}{2019}).
\newblock \eprint{1910.11858}.

\bibitem{Su2021}
\bibinfo{author}{Su, X.} \emph{et~al.}
\newblock \bibinfo{journal}{\bibinfo{title}{Prioritized architecture sampling
  with monto-carlo tree search}}.
\newblock {\emph{\JournalTitle{CoRR}}}
  \textbf{\bibinfo{volume}{abs/2103.11922}} (\bibinfo{year}{2021}).
\newblock \eprint{2103.11922}.

\bibitem{Brabazon2015}
\bibinfo{author}{Brabazon, A.}, \bibinfo{author}{O'Neill, M.} \&
  \bibinfo{author}{McGarraghy, S.}
\newblock \emph{\bibinfo{title}{Natural Computing Algorithms}}
  (\bibinfo{publisher}{Springer Publishing Company, Incorporated},
  \bibinfo{year}{2015}), \bibinfo{edition}{1st} edn.

\bibitem{Kristan2016}
\bibinfo{author}{Kristan, W.~B.}
\newblock \bibinfo{journal}{\bibinfo{title}{Early evolution of neurons}}.
\newblock {\emph{\JournalTitle{Current Biology}}}
  \textbf{\bibinfo{volume}{26}}, \bibinfo{pages}{R949--R954}
  (\bibinfo{year}{2016}).

\bibitem{lanham2023}
\bibinfo{author}{Lanham, M.}
\newblock \emph{\bibinfo{title}{Evolutionary Deep Learning: Genetic algorithms
  and neural networks}} (\bibinfo{publisher}{Manning}, \bibinfo{year}{2023}).

\bibitem{Stanley2019}
\bibinfo{author}{Stanley, K.}, \bibinfo{author}{Clune, J.},
  \bibinfo{author}{Lehman, J.} \& \bibinfo{author}{Miikkulainen, R.}
\newblock \bibinfo{journal}{\bibinfo{title}{Designing neural networks through
  neuroevolution}}.
\newblock {\emph{\JournalTitle{Nature Machine Intelligence}}}
  \textbf{\bibinfo{volume}{1}}, \bibinfo{pages}{24–35},
  \doiprefix\url{10.1038/s42256-018-0006-z} (\bibinfo{year}{2019}).

\bibitem{Fekia2011}
\bibinfo{author}{Fekia\u{c}, J.}, \bibinfo{author}{Zelinka, I.} \&
  \bibinfo{author}{Burguillo, J.~C.}
\newblock \bibinfo{title}{A review of methods for encoding neural network
  topologies in evolutionary computation}.
\newblock In \emph{\bibinfo{booktitle}{European Conference on Modelling and
  Simulation}} (\bibinfo{year}{2011}).

\bibitem{Hara2003}
\bibinfo{author}{Hara, F.} \& \bibinfo{author}{Pfeifer, R.}
\newblock \emph{\bibinfo{title}{Morpho-functional Machines: The New Species:
  Designing Embodied Intelligence}} (\bibinfo{year}{2003}).

\bibitem{Bordin2023}
\bibinfo{author}{Bordin, N.} \emph{et~al.}
\newblock \bibinfo{journal}{\bibinfo{title}{Novel machine learning approaches
  revolutionize protein knowledge}}.
\newblock {\emph{\JournalTitle{Trends in Biochemical Sciences}}}
  \textbf{\bibinfo{volume}{48}}, \bibinfo{pages}{345--359},
  \doiprefix\url{https://doi.org/10.1016/j.tibs.2022.11.001}
  (\bibinfo{year}{2023}).

\bibitem{Matos2009}
\bibinfo{author}{Matos, A.}, \bibinfo{author}{Suzuki, R.} \&
  \bibinfo{author}{Arita, T.}
\newblock \bibinfo{journal}{\bibinfo{title}{Heterochrony and artificial
  embryogeny: A method for analyzing artificial embryogenies based on
  developmental dynamics}}.
\newblock {\emph{\JournalTitle{Artif. Life}}} \textbf{\bibinfo{volume}{15}},
  \bibinfo{pages}{131–160}, \doiprefix\url{10.1162/artl.2009.15.2.15200}
  (\bibinfo{year}{2009}).

\bibitem{Salzberg2018O}
\bibinfo{author}{Salzberg, S.~L.}
\newblock \bibinfo{journal}{\bibinfo{title}{Open questions: How many genes do
  we have?}}
\newblock {\emph{\JournalTitle{BMC Biology}}} \textbf{\bibinfo{volume}{16}}
  (\bibinfo{year}{2018}).

\bibitem{HerculanoHouzel2009}
\bibinfo{author}{Herculano‐Houzel, S.}
\newblock \bibinfo{journal}{\bibinfo{title}{The human brain in numbers: A
  linearly scaled-up primate brain}}.
\newblock {\emph{\JournalTitle{Frontiers in Human Neuroscience}}}
  \textbf{\bibinfo{volume}{3}} (\bibinfo{year}{2009}).

\bibitem{Lim2014}
\bibinfo{author}{Lim, W.}, \bibinfo{author}{Mayer, B.} \&
  \bibinfo{author}{Pawson, T.}
\newblock \emph{\bibinfo{title}{Cell Signaling}} (\bibinfo{publisher}{CRC
  Press}, \bibinfo{year}{2014}).

\bibitem{Tang2022}
\bibinfo{author}{Tang, Q.-Y.}, \bibinfo{author}{Ren, W.},
  \bibinfo{author}{Wang, J.} \& \bibinfo{author}{Kaneko, K.}
\newblock \bibinfo{journal}{\bibinfo{title}{{The Statistical Trends of Protein
  Evolution: A Lesson from AlphaFold Database}}}.
\newblock {\emph{\JournalTitle{Molecular Biology and Evolution}}}
  \textbf{\bibinfo{volume}{39}}, \bibinfo{pages}{msac197},
  \doiprefix\url{10.1093/molbev/msac197} (\bibinfo{year}{2022}).
\newblock
  \eprint{https://academic.oup.com/mbe/article-pdf/39/10/msac197/46424008/msac197.pdf}.

\bibitem{Kenneth2002}
\bibinfo{author}{Stanley, K.~O.} \& \bibinfo{author}{Miikkulainen, R.}
\newblock \bibinfo{journal}{\bibinfo{title}{Evolving neural networks through
  augmenting topologies}}.
\newblock {\emph{\JournalTitle{Evolutionary Computation}}}
  \textbf{\bibinfo{volume}{10}}, \bibinfo{pages}{99--127},
  \doiprefix\url{10.1162/106365602320169811} (\bibinfo{year}{2002}).

\bibitem{Jumper2021}
\bibinfo{author}{Jumper, J.} \emph{et~al.}
\newblock \bibinfo{journal}{\bibinfo{title}{Highly accurate protein structure
  prediction with alphafold}}.
\newblock {\emph{\JournalTitle{Nature}}} \textbf{\bibinfo{volume}{596}},
  \bibinfo{pages}{1--11}, \doiprefix\url{10.1038/s41586-021-03819-2}
  (\bibinfo{year}{2021}).

\bibitem{Abramson2024}
\bibinfo{author}{Abramson, J.} \emph{et~al.}
\newblock \bibinfo{journal}{\bibinfo{title}{Accurate structure prediction of
  biomolecular interactions with alphafold{\thinspace}3}}.
\newblock {\emph{\JournalTitle{Nature}}}
  \doiprefix\url{10.1038/s41586-024-07487-w} (\bibinfo{year}{2024}).

\bibitem{Gao2022}
\bibinfo{author}{Gao, M.}, \bibinfo{author}{Nakajima~An, D.},
  \bibinfo{author}{Parks, J.~M.} \& \bibinfo{author}{Skolnick, J.}
\newblock \bibinfo{journal}{\bibinfo{title}{Af2complex predicts direct physical
  interactions in multimeric proteins with deep learning}}.
\newblock {\emph{\JournalTitle{Nature Communications}}}
  \textbf{\bibinfo{volume}{13}}, \doiprefix\url{10.1038/s41467-022-29394-2}
  (\bibinfo{year}{2022}).

\bibitem{Jinhyuk2019}
\bibinfo{author}{Lee, J.} \emph{et~al.}
\newblock \bibinfo{journal}{\bibinfo{title}{{BioBERT: a pre-trained biomedical
  language representation model for biomedical text mining}}}.
\newblock {\emph{\JournalTitle{Bioinformatics}}} \textbf{\bibinfo{volume}{36}},
  \bibinfo{pages}{1234--1240}, \doiprefix\url{10.1093/bioinformatics/btz682}
  (\bibinfo{year}{2019}).
\newblock
  \eprint{https://academic.oup.com/bioinformatics/article-pdf/36/4/1234/48983216/bioinformatics\_36\_4\_1234.pdf}.

\bibitem{Mirdita2022}
\bibinfo{author}{Mirdita, M.} \emph{et~al.}
\newblock \bibinfo{journal}{\bibinfo{title}{Colabfold: making protein folding
  accessible to all}}.
\newblock {\emph{\JournalTitle{Nature Methods}}} \textbf{\bibinfo{volume}{19}},
  \bibinfo{pages}{679--682}, \doiprefix\url{10.1038/s41592-022-01488-1}
  (\bibinfo{year}{2022}).

\bibitem{Tieri2013}
\bibinfo{author}{Tieri, P.} \& \bibinfo{author}{Nardini, C.}
\newblock \bibinfo{journal}{\bibinfo{title}{Signalling pathway database
  usability: lessons learned}}.
\newblock {\emph{\JournalTitle{Mol. BioSyst.}}} \textbf{\bibinfo{volume}{9}},
  \bibinfo{pages}{2401--2407}, \doiprefix\url{10.1039/C3MB70242A}
  (\bibinfo{year}{2013}).

\bibitem{Loiola2007}
\bibinfo{author}{Loiola, E.~M.}, \bibinfo{author}{{de Abreu}, N. M.~M.},
  \bibinfo{author}{Boaventura-Netto, P.~O.}, \bibinfo{author}{Hahn, P.} \&
  \bibinfo{author}{Querido, T.}
\newblock \bibinfo{journal}{\bibinfo{title}{A survey for the quadratic
  assignment problem}}.
\newblock {\emph{\JournalTitle{European Journal of Operational Research}}}
  \textbf{\bibinfo{volume}{176}}, \bibinfo{pages}{657--690},
  \doiprefix\url{https://doi.org/10.1016/j.ejor.2005.09.032}
  (\bibinfo{year}{2007}).

\bibitem{Albert2002}
\bibinfo{author}{Albert, R.} \& \bibinfo{author}{Barab\'asi, A.-L.}
\newblock \bibinfo{journal}{\bibinfo{title}{Statistical mechanics of complex
  networks}}.
\newblock {\emph{\JournalTitle{Rev. Mod. Phys.}}}
  \textbf{\bibinfo{volume}{74}}, \bibinfo{pages}{47--97},
  \doiprefix\url{10.1103/RevModPhys.74.47} (\bibinfo{year}{2002}).

\bibitem{Leskovec2010}
\bibinfo{author}{Leskovec, J.}, \bibinfo{author}{Chakrabarti, D.},
  \bibinfo{author}{Kleinberg, J.}, \bibinfo{author}{Faloutsos, C.} \&
  \bibinfo{author}{Ghahramani, Z.}
\newblock \bibinfo{journal}{\bibinfo{title}{Kronecker graphs: An approach to
  modeling networks}}.
\newblock {\emph{\JournalTitle{J. Mach. Learn. Res.}}}
  \textbf{\bibinfo{volume}{11}}, \bibinfo{pages}{985–1042}
  (\bibinfo{year}{2010}).

\bibitem{Minsu2010}
\bibinfo{author}{Cho, M.}, \bibinfo{author}{Lee, J.} \& \bibinfo{author}{Lee,
  K.~M.}
\newblock \bibinfo{title}{Reweighted random walks for graph matching}.
\newblock In \bibinfo{editor}{Daniilidis, K.}, \bibinfo{editor}{Maragos, P.} \&
  \bibinfo{editor}{Paragios, N.} (eds.) \emph{\bibinfo{booktitle}{Computer
  Vision -- ECCV 2010}}, \bibinfo{pages}{492--505}
  (\bibinfo{publisher}{Springer Berlin Heidelberg}, \bibinfo{address}{Berlin,
  Heidelberg}, \bibinfo{year}{2010}).

\bibitem{Yan2015}
\bibinfo{author}{Yan, J.} \emph{et~al.}
\newblock \bibinfo{journal}{\bibinfo{title}{Discrete hyper-graph matching}}.
\newblock {\emph{\JournalTitle{2015 IEEE Conference on Computer Vision and
  Pattern Recognition (CVPR)}}} \bibinfo{pages}{1520--1528}
  (\bibinfo{year}{2015}).

\bibitem{Ngoc2015}
\bibinfo{author}{Ngoc, Q.~N.}, \bibinfo{author}{Gautier, A.} \&
  \bibinfo{author}{Hein, M.}
\newblock \bibinfo{journal}{\bibinfo{title}{A flexible tensor block coordinate
  ascent scheme for hypergraph matching}}.
\newblock {\emph{\JournalTitle{2015 IEEE Conference on Computer Vision and
  Pattern Recognition (CVPR)}}} \bibinfo{pages}{5270--5278}
  (\bibinfo{year}{2015}).

\bibitem{Cai2017}
\bibinfo{author}{Cai, H.}, \bibinfo{author}{Zheng, V.~W.} \&
  \bibinfo{author}{Chang, K. C.-C.}
\newblock \bibinfo{journal}{\bibinfo{title}{A comprehensive survey of graph
  embedding: Problems, techniques, and applications}}.
\newblock {\emph{\JournalTitle{IEEE Transactions on Knowledge and Data
  Engineering}}} \textbf{\bibinfo{volume}{30}}, \bibinfo{pages}{1616--1637}
  (\bibinfo{year}{2017}).

\bibitem{Qiao2023}
\bibinfo{author}{Qiao, Y.} \& \bibinfo{author}{Gallagher, M.}
\newblock \bibinfo{title}{Modularity based linkage model for neuroevolution}.
\newblock In \bibinfo{editor}{Silva, S.} \& \bibinfo{editor}{Paquete, L.}
  (eds.) \emph{\bibinfo{booktitle}{Companion Proceedings of the Conference on
  Genetic and Evolutionary Computation, {GECCO} 2023, Companion Volume, Lisbon,
  Portugal, July 15-19, 2023}}, \bibinfo{pages}{675--678},
  \doiprefix\url{10.1145/3583133.3590648} (\bibinfo{publisher}{{ACM}},
  \bibinfo{year}{2023}).

\bibitem{Lynch2016}
\bibinfo{author}{Lynch, M.} \emph{et~al.}
\newblock \bibinfo{journal}{\bibinfo{title}{Genetic drift, selection and the
  evolution of the mutation rate}}.
\newblock {\emph{\JournalTitle{Nature Reviews Genetics}}}
  \textbf{\bibinfo{volume}{17}}, \bibinfo{pages}{704--714}
  (\bibinfo{year}{2016}).

\bibitem{Kaessmann2010}
\bibinfo{author}{Kaessmann, H.}
\newblock \bibinfo{journal}{\bibinfo{title}{Origins, evolution, and phenotypic
  impact of new genes.}}
\newblock {\emph{\JournalTitle{Genome research}}} \textbf{\bibinfo{volume}{20
  10}}, \bibinfo{pages}{1313--26} (\bibinfo{year}{2010}).

\bibitem{bucciarelli2022}
\bibinfo{author}{Bucciarelli, G.}, \bibinfo{author}{Alsalek, F.},
  \bibinfo{author}{Kats, L.}, \bibinfo{author}{Green, D.} \&
  \bibinfo{author}{Shaffer, H.}
\newblock \bibinfo{journal}{\bibinfo{title}{Toxic relationships and arms-race
  coevolution revisited}}.
\newblock {\emph{\JournalTitle{Annual Review of Animal Biosciences}}}
  \textbf{\bibinfo{volume}{10}}, \bibinfo{pages}{63--80},
  \doiprefix\url{10.1146/annurev-animal-013120-024716} (\bibinfo{year}{2022}).
\newblock \bibinfo{note}{PMID: 35167315},
  \eprint{https://doi.org/10.1146/annurev-animal-013120-024716}.

\bibitem{Lazzaro2023}
\bibinfo{author}{Lazzaro, D.} \emph{et~al.}
\newblock \emph{\bibinfo{title}{Minimizing Energy Consumption of Deep Learning
  Models by Energy-Aware Training}}, \bibinfo{pages}{515--526}
  (\bibinfo{year}{2023}).

\bibitem{Lechner2020}
\bibinfo{author}{Lechner, M.} \emph{et~al.}
\newblock \bibinfo{journal}{\bibinfo{title}{Neural circuit policies enabling
  auditable autonomy}}.
\newblock {\emph{\JournalTitle{Nature Machine Intelligence}}}
  \textbf{\bibinfo{volume}{2}}, \bibinfo{pages}{642 -- 652}
  (\bibinfo{year}{2020}).

\bibitem{Hoefer2024}
\bibinfo{author}{H\"{o}fer, T.}, \bibinfo{author}{Bierwirth, S.} \&
  \bibinfo{author}{Madlener, R.}
\newblock \bibinfo{title}{C15 ‐ energie mehrverbrauch in rechenzentren bei
  einführung des 5g standards} (\bibinfo{year}{2024}).

\bibitem{Patterson2021}
\bibinfo{author}{Patterson, D.~A.} \emph{et~al.}
\newblock \bibinfo{journal}{\bibinfo{title}{Carbon emissions and large neural
  network training}}.
\newblock {\emph{\JournalTitle{ArXiv}}}
  \textbf{\bibinfo{volume}{abs/2104.10350}} (\bibinfo{year}{2021}).

\bibitem{Strubell2019}
\bibinfo{author}{Strubell, E.}, \bibinfo{author}{Ganesh, A.} \&
  \bibinfo{author}{McCallum, A.}
\newblock \bibinfo{journal}{\bibinfo{title}{Energy and policy considerations
  for deep learning in nlp}}.
\newblock {\emph{\JournalTitle{ArXiv}}}
  \textbf{\bibinfo{volume}{abs/1906.02243}} (\bibinfo{year}{2019}).

\bibitem{GreeningICT}
\bibinfo{title}{Greening ict}.
\newblock
  \bibinfo{howpublished}{\url{{https://ec.europa.eu/assets/rtd/srip/chapter4.3.html}}}.
\newblock \bibinfo{note}{Accessed: 2024-03-10}.

\bibitem{CyberSeek2024}
\bibinfo{title}{Cyberseek}.
\newblock \bibinfo{howpublished}{\url{https://www.cyberseek.org/heatmap.html}}.
\newblock \bibinfo{note}{Accessed: 2024-03-10}.

\bibitem{AICybersecurity}
\bibinfo{title}{Artificial intelligence in cybersecurity market share,
  forecast}.
\newblock
  \bibinfo{howpublished}{\url{https://www.marketsandmarkets.com/Market-Reports/artificial-intelligence-ai-cyber-security-market-220634996.html}}.
\newblock \bibinfo{note}{Accessed: 2024-03-10}.

\bibitem{Lewis2018}
\bibinfo{author}{Lewis, J.}
\newblock \bibinfo{title}{Economic impact of cybercrime, no slowing down}.
\newblock \bibinfo{type}{Tech. Rep.}, \bibinfo{institution}{McAfee}
  (\bibinfo{year}{2018}).

\bibitem{economicimpact}
\bibinfo{title}{Economic impact of cybercrime on business predicted to reach
  \$10.5 trillion by 2025: Cybersecurity ventures}.
\newblock
  \bibinfo{howpublished}{\url{https://www.reinsurancene.ws/economic-impact-of-cybercrime-on-business-predicted-to-reach-10-5-trillion-by-2025-cybersecurity-ventures/}}.
\newblock \bibinfo{note}{Accessed: 2024-03-10}.

\bibitem{Lippert2021}
\bibinfo{author}{Lippert, K.~J.} \& \bibinfo{author}{Cloutier, R.}
\newblock \bibinfo{journal}{\bibinfo{title}{Cyberspace: A digital ecosystem}}.
\newblock {\emph{\JournalTitle{Syst.}}} \textbf{\bibinfo{volume}{9}},
  \bibinfo{pages}{48} (\bibinfo{year}{2021}).

\bibitem{Bacar2011}
\bibinfo{author}{Baca{\"e}r, N.}
\newblock \bibinfo{title}{Lotka, volterra and the predator–prey system
  (1920–1926)}.
\newblock \doiprefix\url{10.1007/978-0-85729-115-8_13} (\bibinfo{year}{2011}).

\bibitem{Guthikonda2017}
\bibinfo{author}{Guthikonda, A.}, \bibinfo{author}{Al-Shaer, E.},
  \bibinfo{author}{Farooq, A.} \& \bibinfo{author}{Raja, M.~Y.}
\newblock \bibinfo{title}{Bio-inspired innovations in cyber security}.
\newblock \bibinfo{pages}{105--109}, \doiprefix\url{10.1109/HONET.2017.8102212}
  (\bibinfo{year}{2017}).

\end{thebibliography}

\end{document}